 \documentclass[tablecaption=bottom,wcp]{jmlr} 

\usepackage{booktabs}
\usepackage[load-configurations=version-1]{siunitx} 

\theorembodyfont{\upshape}
\theoremheaderfont{\scshape}
\theorempostheader{:}
\theoremsep{\newline}

\jmlrvolume{1}
\jmlryear{2020}
\jmlrsubmitted{submission date}
\jmlrpublished{publication date}
\jmlrworkshop{Demonstrations Track @NeurIPS2019} 

\title[Empathic AI Painter]{Empathic AI Painter: \titlebreak A Computational Creativity System with Embodied Conversational Interaction}

  \author{\Name{{\"O}zge Nilay Yal{\c{c}}in} \Email{oyalcin@sfu.ca}\\
   \Name{Nouf Abukhodair} \Email{nouf\_abukhodair\_2@sfu.ca}\\
   \Name{Steve DiPaola} \Email{sdipaola@sfu.ca}\\
  \addr School of Interactive Arts and Technology,
  Simon Fraser University,
  Surrey, BC, CANADA}

\editor{Editor's name}

\begin{document}

\maketitle

\begin{abstract}
There is a growing recognition that artists use valuable ways to understand and work with cognitive and perceptual mechanisms to convey desired experiences and narrative in their created artworks \citep{dipaola2010rembrandt, zeki2001artistic}. 
This paper documents our attempt to computationally model the creative process of a portrait painter, who relies on understanding human traits (i.e., personality and emotions) to inform their art. Our system includes an empathic conversational interaction component to capture the dominant personality category of the user and a generative AI Portraiture system that uses this categorization to create a personalized stylization of the user's portrait. This paper includes the description of our systems and the real-time interaction results obtained during the demonstration session of the NeurIPS 2019 Conference.
\end{abstract}

\section{Introduction}
\label{sec:intro}

Representation of human traits has always been of great value in the creation of artworks. Even though science and art seem to have very different toolsets and goals, these differences could be seen as blurred where an artist seems to follow scientific knowledge to inform their craft and where science studies art to understand the human condition. Using well-established insights from psychological research on personality, emotion, and other human traits to inform, create and critique artwork is not a new idea \citep{mccrae2012five}. 

Traditional portrait artists use knowledge on human vision and perception to create a painterly portrait of a live or photographed sitter \citep{dipaola2010rembrandt}. To create a portrait, a human portrait painter would not only set up the room/lighting, position the sitter, interview the sitter to understand/capture their physical and mental characteristics, but also try to convey their painting style in the trajectory of their painting career as well as strive for some personal and universal (e.g., emotional and cultural concepts) truth of the current world they live in. An artist has a palette of choices of themes, brush style, colour plan, edge and line plan, abstraction style, and emotional narrative at their disposal to create the final painting that balances these goals.

The discipline of computational creativity and generative art opens up new possibilities for modeling scientific knowledge to emulate this process and advance our understanding of human creativity further. Our research uses a host of artificial intelligence techniques as an attempt to begin to understand and emulate this creative process. The system we created, the Empathic AI Painter, is aimed to examine new ways of balancing or blending different aesthetic, conceptual, abstraction concerns at a semantic level. We showcased our work at the demo session of the NeurIPS 2019 Conference in Vancouver, Canada. This paper includes a detailed description of the research, the process and the demonstration, followed by some discussion on the results obtained and future work.

\section{System Description}
\label{sec:sys}
The Empathic Painter System is designed to emulate the interaction of a live portrait artist with a human (the ‘sitter’ in art terms). It attempts to understand the traits (i.e., emotion and personality) to emphatically create a unique portrait of the sitter through selecting the right color palette, style, and abstraction techniques that match the sitter's emotion and personality traits. Our system incorporates two distinct components implemented as a two-stage process: in the first step, it aims to capture the characteristics of the sitter; second, using these traits to generate an artistic representation of their portrait during the second stage of the process (see \figureref{fig:System}).

\begin{figure}[htbp]
	\floatconts
	{fig:System}
	{\caption{The two components of the system: conversational interaction and generative portrait stylization. The Big-5 categorical mapping between these components is used to create a personality-based stylized portrait of the sitter.}}
	{\includegraphics[width=0.95\linewidth]{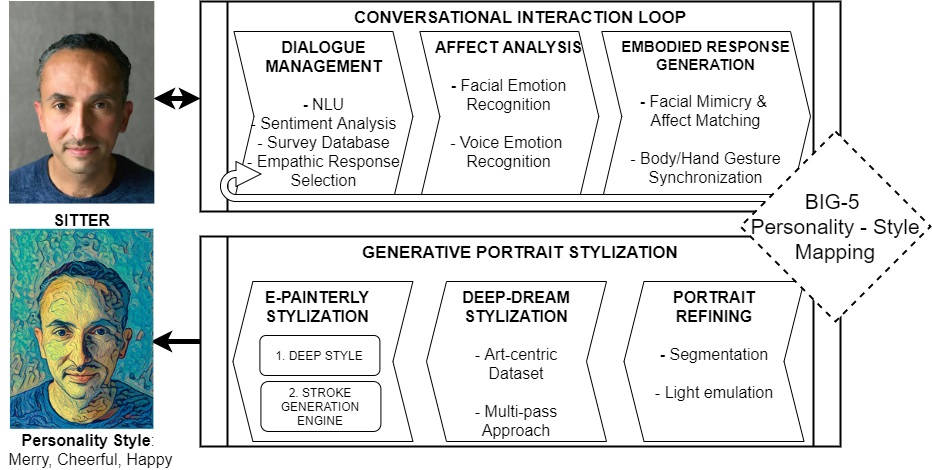}}
\end{figure}

The first stage, capturing the personality-based characteristics of the sitter, is achieved during the conversational interaction of the sitter with our embodied conversational agent that uses empathic interaction methods. We are using the M-Path conversational agent system \citep{yalccin2020empathy} that was previously developed by our research group. For this demonstration, we modified the M-Path system to perform an interview based on the Big-5 personality questionnaire (see \sectionref{sec:big5} and \sectionref{sec:eca}) to categorize the sitter in one of the well-established personality dimensions. This information is used to map the personality traits to a particular artistic style. This mapping is passed to our Generative AI Portrait Stylization system at the second stage, to generate an artistic portrait of the sitter (see \sectionref{sec:genart}). 

\begin{figure}[htbp]
	\floatconts
	{fig:ECA}
	{\caption{An example setup of the interaction with the M-Path system, where the 3D embodied conversational agent conducts Big-5 questionnaire with the participant using a webcam and microphone.}}
	{\includegraphics[width=0.8\linewidth]{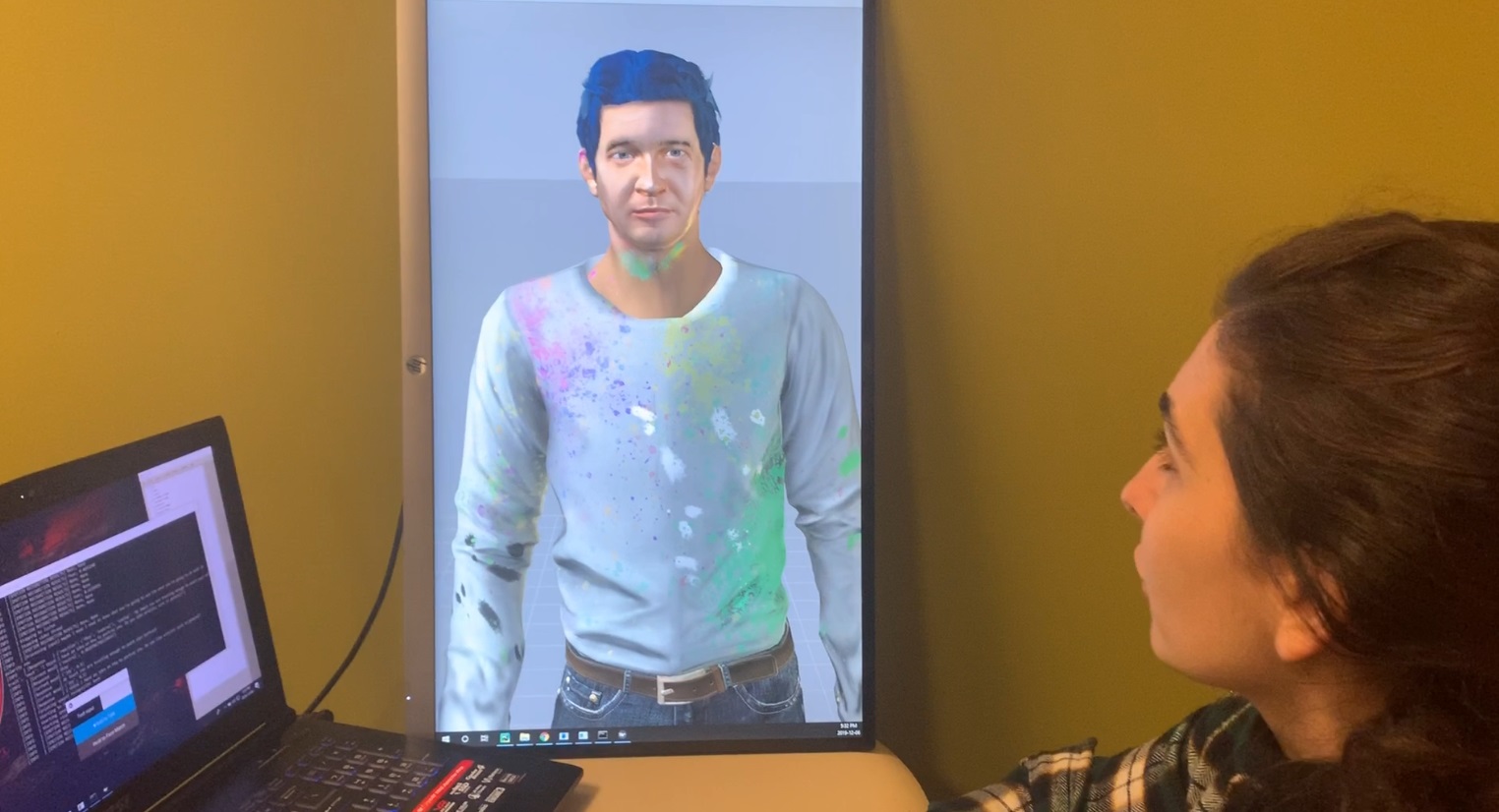}}
\end{figure}

The interaction scenario used during the demonstration included several steps. First, the source portrait of the sitter is taken in a controlled lighting condition, and a unique ID is assigned while obtaining consent for participation and usage of the portrait. Then, the sitter is informed about the M-Path system and given instructions on how to interact with the conversational agent. The sitter then initiates the interaction until a successful conversation is achieved, and the conversational agent informs the sitter that the interaction is over. M-Path system uses the information gathered during the interaction to classify the personality dimension of the sitter as a category. This category is then passed to the Generative AI Portraiture system to generate a personalized portrait style of the sitter. The generated portraits are displayed on a large monitor for all the participants and the demonstration crowd to see and evaluate. \figureref{fig:ECA} shows an example interaction setting with the M-Path system. Details on both stages and the mapping process are described in the following sections.

\subsection{Big-5 Personality Mapping}
\label{sec:big5}
The five-factor model of personality (FFM), also referred to as the "Big-5 Personality Model," is constructed as an empirical generalization to capture the covariation of widely-accepted personality traits between individuals \citep{mccrae1989structure}. It is constructed as an attempt to categorize the personality variations along five dimensions: extraversion, openness, conscientiousness, neuroticism, and agreeableness.

Each personality dimension describes a broad domain of psychological functioning that is composed from a set of more specific and narrow traits \cite{zhao2006big, barrick1991big}. Extraversion describes the extent to which people are assertive, dominant, energetic, active, talkative, and enthusiastic. Openness is a personality dimension that characterizes someone who is intellectually curious and tends to seek new experiences and explore novel ideas, which can be described as creative, innovative, imaginative, reflective, untraditional, cultured, curious, original, broad-minded, intelligent, and artistically sensitive. Conscientiousness indicates an individual’s degree of organization, persistence, hard work, and motivation in the pursuit of goal accomplishment. People on the high end of this dimension are thought to be organized, plan oriented, and determined. Neuroticism represents individual differences in adjustment and emotional stability, which is also called Emotional Stability, Stability, or Emotionality. Individuals with high Neuroticism scores tend to experience a number of negative emotions including anxiety, hostility, depression, self-consciousness, impulsiveness, vulnerability, anger, embarrassment, and worrisome. Lastly, agreeableness has been interpreted as likability, friendliness, social conformity or compliance [1]. Individuals with high Agreeableness scores can be characterized as trusting, forgiving, caring, altruistic, gullible, courteous, flexible, good-natured, cooperative, soft-hearted, and tolerant.

Several independent researchers used factor analysis of verbal descriptors of human behavior to define this model \citep{john1999big}. The instrument used in this paper is based on the shortened version of the Revised NEO Personality Inventory (NEO-PI-R) by \citet{costa2008revised}. The original questionnaire consists of 120 statements, 25 statements for each dimension that takes about 45 minutes to finish. For the online demonstration purposes, we only used one statement per each dimension where the whole conversational interaction takes less than 5 minutes to complete. Each question is further modified to fit the conversation setup in the conference environment (see \tableref{tab:questions}). 

\begin{table}[hbtp]
	\floatconts
	{tab:questions}
	{\caption{The questions used for the personality dimensions.}}
	{\begin{tabular}{ll}
			\toprule
			\bfseries Dimension & \bfseries Question\\
			\midrule
			Openness & How do you like the conference so far, is it interesting to you?\\
			Conscientiousness & Don't you think the conferences are always a bit chaotic?\\
			Extraversion & Do you normally talk and interact with a lot of people?\\
			Agreeableness & How about agents? Do you trust me in sharing how you feel?\\
			Neuroticism & How do you feel about your portrait being displayed on the screen?\\
			\bottomrule
	\end{tabular}}
\end{table}

The answers to these questions are then evaluated based on its polarity (i.e., positive, neutral, negative) and mapped onto two-factor dimensions for personality adjectives. The mapping model we used in this paper is the Abridged Big Five Circumplex Model (AB5C), where the facets of the Big Five dimensions are mapped as combinations of two factors \citep{hofstee1992integration}. AB5C mapping includes descriptive personality terms for each of the 90 resulting combinations, where the most distinctive trait of an individual is used to select the column, and the next most distinctive trait selects the row. The traits could be either positive or negative.

The mapping from our Big-5 traits to our Generative AI portrait styles was given to art experts in our research group, who independently matched the styles into the Big-5 categories and came to an agreement. We planned a study and mechanism to refine and verify these results using both an art and psychology students. However, given the compressed timescale of the demonstration, we were not able to schedule for the demo. We are looking to complete such a study in the future.



\subsection{Empathic Conversational Avatar}
\label{sec:eca}
The initial point of interaction for our system is the empathic conversational agent, M-Path \citep{yalccin2020empathy}, which was developed using a framework that is based on a computational model of empathy \citep{yalcin2018computational}. M-Path is implemented in an embodied human-like avatar that is capable of initiating and maintaining an emotional conversation, based on the predetermined goal of the dialogue\footnote{The conversational system is shared for open-source usage and can be reached at \url{https://github.com/onyalcin/M-PATH}}. The interaction scenario includes a face-to-face conversation with a human interaction partner, similar to a video-conference setting with audio-visual input and output modalities. The agent processes the real-time inputs from the interaction partner in terms of their affective and linguistic properties, to be able to generate empathic verbal and non-verbal response behavior (see \figureref{fig:mpath}). The overall goal of the dialogue is to complete the modified Big-5 questionnaire, to be able to assign a personality category to the conversation partner and send this information to be further processed in the generative art system.

\begin{figure}[htbp]
	\floatconts
	{fig:mpath}
	{\caption{Our Empathic Conversational Agent system and example captured avatar behavior.}}
	{\includegraphics[width=0.95\linewidth]{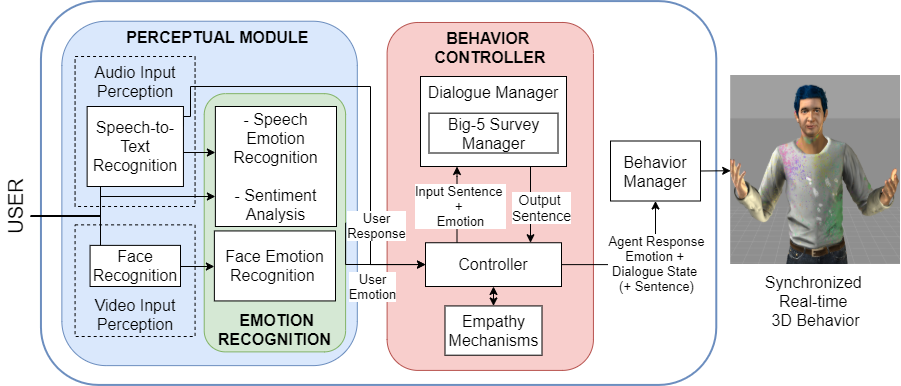}}
\end{figure}

In the following sections, we will give a detailed description of the empathic conversational system based on its input processing, decision making and output generation capabilities. These mechanisms are controlled by three distinct modules within the system: perceptual module, behavior controller and behavior manager. 

\paragraph{Perceptual Module}
M-Path system processes the participants' input in its perceptual module. This module gathers the audio and video input signals via a microphone and a video camera when the conversation partner is speaking. During the demo session, this process was triggered with the aid of the push-to-talk system, where users provide their input via pressing and holding a button. When the user starts speaking using this mechanism, M-Path enters to the “listening” state. During this state, the speech and facial expressions are processed in real-time for speech and emotional understanding. 

The video input is processed in the facial emotion recognition module, with an OpenCV face-recognition algorithm \citep{opencv_library} to detect the face. After detecting the face, the emotions are recognized in 6 basic emotion categories (i.e., anger, disgust, fear, joy, sadness, surprise and contempt), which is categorized by a CNN model trained over CK+ Dataset \citep{lucey2010extended}. The speech input from the microphone is first sent to the speech-to-text module, which uses Google Speech-to-Text (STT) service to get streaming speech recognition \citep{googlespeech}. The sentiment analysis component uses the text received from the STT service to evaluate the response in terms of polarity (positive-neutral-negative) value of the text. In the live-demo, we used the SO-CAL Sentiment Analyzer \citep{taboada2011lexicon}, which was re-trained over the NRC-Canada lexicon \citep{MohammadKZ2013}. The text of the speech is later sent to the decision-making component to generate conversational responses. 

This processing continues until the speech of the conversation partner is finished, which concludes the listening state. This processed information is then sent to the decision-making component as the agent enters the “thinking” state.

\paragraph{Behavior Controller}
The behavior controller module is responsible for generating empathic and goal-directed verbal and non-verbal responses during all the cycles of the conversation: listening, thinking, and speaking. This is achieved by processing the user response and emotion information, both of which were perceived during the listening act. The conversation cycle starts with the user’s initiation with a greeting and ends after the agent receives satisfactory responses to the questions of the personality survey. The listening, thinking and speaking state of the agent is looped in sequence until a successful categorization of the user is reached.
 
During the listening stage, the agent shows non-verbal affect matching response and backchanneling behavior. Affect matching is created with a facial expression that matches the user's facial expressions in real-time, which are chosen by the processes of empathy mechanisms. Backchanneling is created with nodding behavior that is triggered by the detection of the pauses while conversing with the user. Both of these behaviors are combined to generate natural and empathic listening behavior. Previous work showed that the empathic responses generated during both the listening and speaking stages were indeed increasing the perception of the agent as empathic \citep{yalcin2019evaluating}.  

After the conversation with the participant is concluded, the final result received from the STT engine is passed to the Dialogue Manager (DM) along with the overall sentiment of the user, and ultimately sent to the Empathy Mechanisms (EM) component. The goal of the DM is to complete the modified Big-5 personality questionnaire in order to assign a personality category. The goal of the EM is to make sure that the DM generates empathic responses while reaching its own goal. DM collects the appropriate emotional response decided by the EM to create an emotionally appropriate verbal reaction to the user's response, which is followed by a survey-related coping response, and finally the next survey question. Our system uses the scikit-learn library \citep{scikit-learn} in Python for the TF-IDF vectorizer model, while using NLTK Lemmatizer \citep{Loper02nltk:the}. The second model is generated by fine-tuning the pre-trained language representation model BERT \citep{devlin2018bert} for the classification of user responses based on sentiment and the Big-5 questionnaire answers.

The answers to the Big-5 questionnaire are collected to select the most dominant personality dimensions of the user based on the polarity and their assigned probability values. The Big-5 mapping mentioned in the \sectionref{sec:big5} is then used to select a category for the user along with its adjectives. This categorization is then sent to the generative art cycle to create a personalized portrait of the user. After each appropriate response is generated by the dialogue manager component, it is then sent to the behavior manager to be performed by the embodied conversational agent during the speaking state.


\paragraph{Behavior Manager}
In order to create natural conversational behavior, M-Path continuously generates non-verbal and/or verbal behaviors in all states of the dialogue. A combination of facial expressions, body gestures, head gestures, posture and lip movements are synchronized with the speech of the agent and sent as a BML (Behavior Markup Language) message to the Smartbody character animation platform \citep{thiebaux2008smartbody} to display the generated behaviors. 



\subsection{Generative AI Portraiture System}
\label{sec:genart}

\begin{figure}[htbp]
	\floatconts
	{fig:subfigex2}
	{\caption{Process flow of our Generative AI Portraiture system from raw source to final portrait.}}
	{%
		\subfigure[Source Photo][c]{\label{fig:orig}%
			\includegraphics[width=0.23\linewidth]{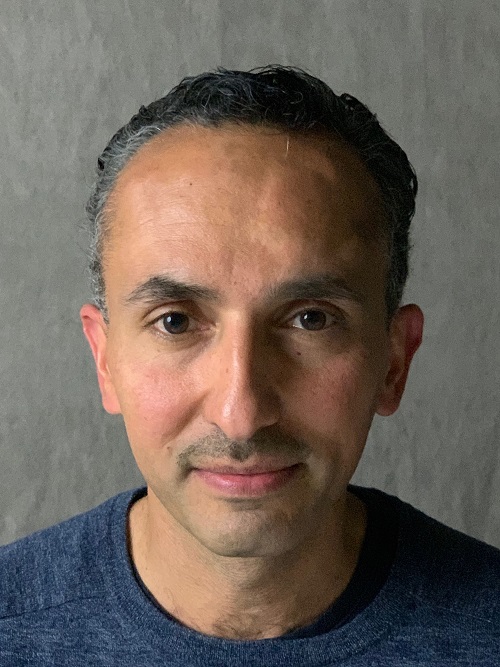}}
		\subfigure[Portrait Refining][c]{\label{fig:pro1}%
			\includegraphics[width=0.23\linewidth]{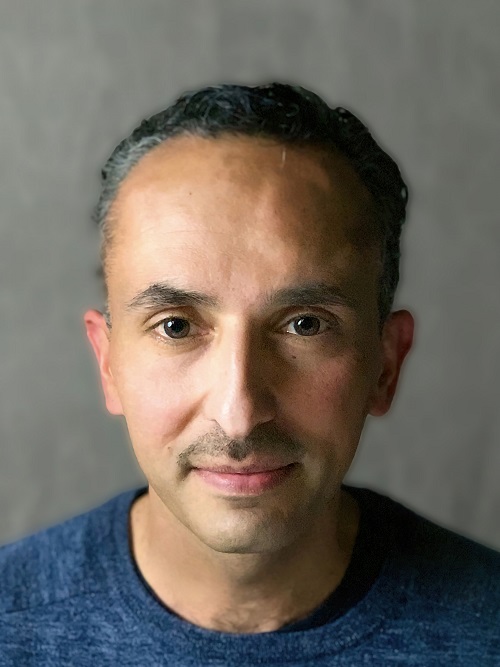}}
		\subfigure[mDD Phase][c]{\label{fig:pro2}%
			\includegraphics[width=0.23\linewidth]{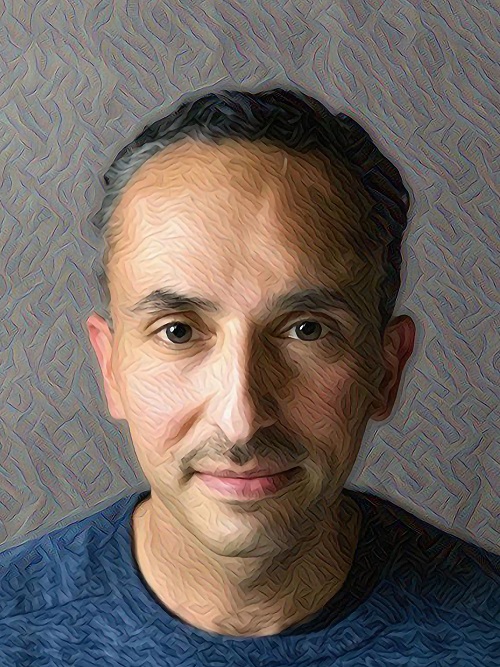}}
		\subfigure[ePainterly Phase][c]{\label{fig:pro3}%
			\includegraphics[width=0.23\linewidth]{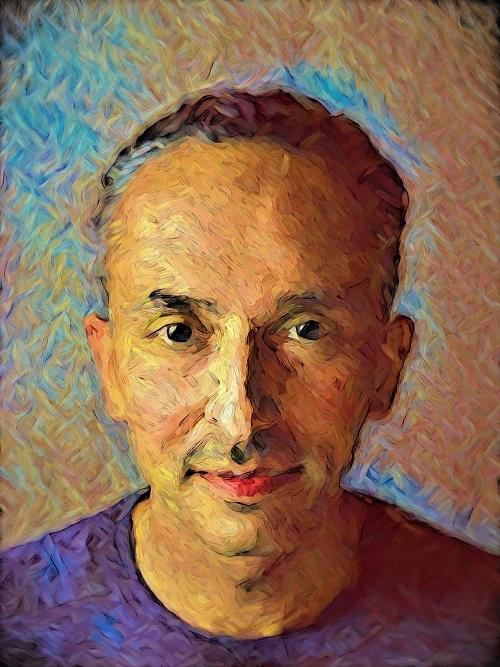}}
	}
\end{figure}

The stylistic rendering of the portraits is generated by the generative art component of our system. Here, the portrait of the sitter goes through three main phases of processing (see \figureref{fig:subfigex2}). In the first phase, the original portrait \subfigref{fig:orig} of the sitter image is pre-processed by using the first AI-processing tool that segments the background from the foreground \cite{bolya2019yolact}, which will be later used to stylized the portrait in a focused way. Then, the light and color balance of the face is achieved to create a Rembrandt lighting effect, where one side of the face is dramatically shown \subfigref{fig:pro1}. 

The next phase uses this image and the personality category as inputs to our modified-Deep Dream (mDD) system with successive passes on the image to create the baseline style \subfigref{fig:pro2}. While most DD systems use pre-trained networks with object recognition data such as ImageNet, we have implemented our own modified DD system (mDD) and train new models with artistic paintings and drawings as training data. We now have amassed a dataset of 160,000 labeled and categorized paintings from 3000 artists for a total size of 67 gigabytes of artistic visual data (one of the largest in an AI research group). However, we have discovered that even with such a large dataset, most artists make under 200 paintings in their lifetime (except Picasso), which might not be rigorous and large enough for an advanced CNN training for art styles. In order to overcome this issue, we developed a method that we call “hierarchical tight style and tile” \citep{dipaola2018informing}.

\begin{figure}[ht]%
	\floatconts
	{fig:pro4}
	{\caption{Example styles created by the system that maps a variety of personality categories shown with the adjectives.}}%
	{\includegraphics[width=0.9\linewidth]{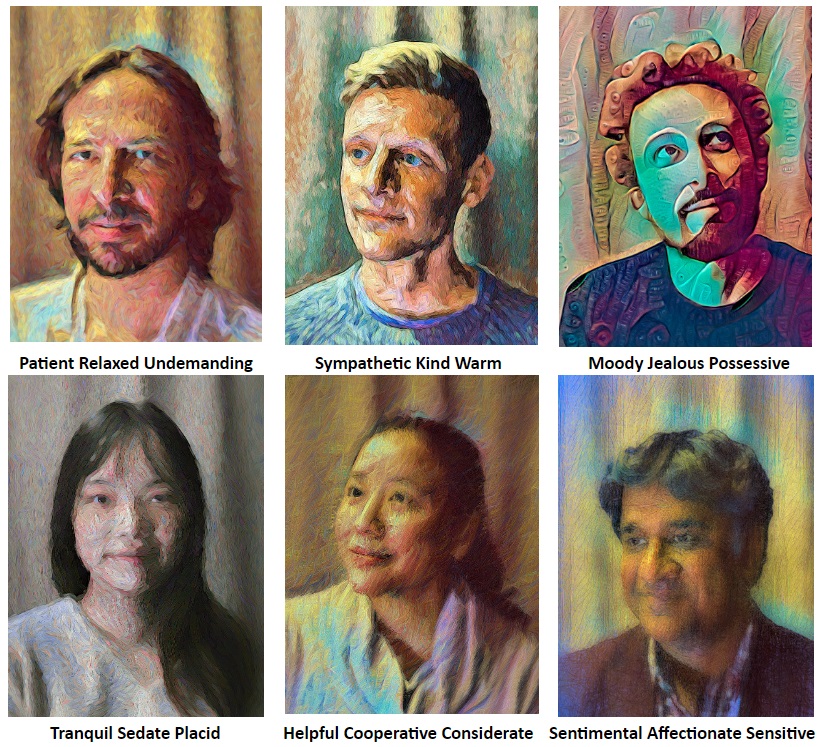}}%
\end{figure}%

In the last phase, the source image created from the previous phase is further refined using the personality category, where our ePainterly system uses a combination of Deep Style techniques as a surface texture manipulator, as well as a series of Non-Photorealistic Rendering (NPR) techniques such as particle systems, color palette manipulation and stroke engine techniques \subfigref{fig:pro3}. This iterative process refines and completes the finished portrait style, and the final result is displayed in an online gallery system for viewing (see \figureref{fig:pro4}). Our ePainterly module is an extension of our former painting system Painterly \citep{dipaola2009incorporating} that models the cognitive processes of artists based on years of research in this area. The NPR subclass of stroke-based rendering is used as the final part of our ePainterly process to realize the internal mDD models with stroke-based output. As it can be seen in \figureref{fig:subfigex2}, the aesthetic advantages of this additional step include reducing noise artifacts from the generated mDD output, cohesive stroke-based clustering, and a better distributed color space. 

\section{Conclusion and Future Work}
\label{sec:math}
The Empathic AI Painter was introduced during the demo session of the NeurIPS 2019 Conference in Vancouver, Canada. A total of 42 participants assigned for testing the system, where 26 of them completed both the portrait-taking and interaction scenario with the system during the 3-hour demonstration session. Each conversational interaction with the M-Path system lasted about 5 minutes. After the demonstration, we evaluated the performance of M-Path interaction individually. On average, 84.72\% of the speech input from the sitter was recognized correctly, where 82.51\% of them had a correct categorization as answers to the personality questionnaire. The 26 participants with a complete interaction showed 17 distinct personality categories. 

A large number of participants contacted us to receive high-resolution versions of their finished portraits. Anecdotally, we heard from the expert conference crowd that the portraits being generated seemed to depict the participants' perceived personality quite well, although we intend to validate this feedback in future studies. We believe our approach has the potential to create better AI generative art systems that can resonate with human traits, which is closer to the creative process of human artists.



We have begun the next phase of this project in three major areas: 1) move beyond the Big-5 model to include affective models for portrait art depiction and evaluation, 2)  plan user studies to determine aspects of cognitive reception of portraits and to validate different user personality models, 3) incorporate our telerobot system to automate the source photo taking and other aspects of our process that have current human intervention.

\bibliography{jmlr-sample}

\end{document}